\documentclass[cameraready]{Interspeech}

\graphicspath{{./figures/}}

\usepackage{booktabs}
\usepackage{multirow}
\usepackage{amsmath}
\usepackage{float}

\title{Tone-Conditioned Curriculum Learning for Low-Resource Bantu Speech Recognition}

\author[affiliation={1},orcid=0009-0000-5713-2002,correspondingauthor]{Kesego}{Mokgosi}
\author[affiliation={2,3}, orcid=0000-0002-6731-6267, equalcontribution]{Vukosi}{Marivate}

\author[affiliation={2}, orcid=0009-0007-5779-1797]{Sitwala}{Mundia}

\author[affiliation={2}, orcid=0009-0003-8567-0977]{Unarine}{Netshifhefhe}
\author[affiliation={2}, orcid=0009-0001-8273-8898]{Tsholofelo Hope}{Mogale}
\author[affiliation={2}, orcid=0000-0002-9754-9547]{Thapelo}{Sindane}

\address{
    $^1$ Technological University Dublin, Ireland
    \\
    $^2$ Data Science for Social Impact, University of Pretoria, South Africa 
    \\
    $^3$ Lelapa AI
}
\email{d23126641@mytudublin.ie}

\usepackage{comment}

\begin{document}

\maketitle

\begin{abstract}
Southern Bantu languages are spoken by over 80 million people, yet current foundation ASR models still produce zero-shot WER above 100\%, which limits practical use in education and public services. We addressed this gap with a tone conditioned curriculum framework for 6 Southern Bantu languages that combined hybrid difficulty scoring, gated adapters driven by tonal statistics and staged curriculum training. We trained on a community corpus and tested transfer to NCHLT to measure robustness beyond matched evaluation. Results revealed clear interactions between architecture and language, with W2V-BERT outperforming Whisper on Nguni languages by 3 to 4 WER points whilst Whisper performed better on Sotho-Tswana languages. W2V-BERT with tone conditioning reached 28.41\% average WER across datasets and 23.79\% on Xitsonga transfer. No single model suited all 6 languages, so deployment should pair model selection per language with validation across corpora.
\end{abstract}

\section{Introduction}

Southern Bantu languages such as isiZulu, isiXhosa, Sesotho, Setswana, Tshivenda and Xitsonga are among the most widely spoken in sub-Saharan Africa~\cite{badenhorst2022nchlt} and hold official status in South Africa. They remain primarily oral, and tone and prosody encode grammatical and cultural distinctions not fully captured in writing. As communities expand language use in education and digital services, limited speech technology remains a barrier~\cite{sutherland2024navajo, reitmaier2022opportunities}. Foundation models such as Whisper~\cite{radford2023robust} and MMS~\cite{pratap2024scaling} yielded zero-shot Word Error Rate (WER) above 100\% and required targeted fine-tuning that reflects the phonology of each language. Bantu languages also pose specific challenges for automatic speech recognition because, unlike Mandarin where tone is largely localised to the syllable, Bantu systems exhibit high tone spreading and phrasal contours that encode lexical meaning and grammatical relations~\cite{zerbian08_interspeech, mixdorff11_interspeech}. Standard orthographies usually omit tone, so models must recover tonal dependencies across morphological boundaries~\cite{mohasi2011acoustic}, which creates a natural difficulty gradient for curriculum learning.

Curriculum scoring based on WER improves Automatic Speech Recognition (ASR) but ignores phonological properties of individual languages~\cite{karakasidis22_interspeech}. A phonologically informed approach is needed because generic fine-tuning overlooks the tonal and morphological contrasts specific to Bantu languages, so we asked whether combining WER difficulty with morphotonal features improved curriculum learning and whether tone conditioned adapters and curriculum scheduling benefited different ASR architectures equally.

Our contributions are as follows:
\begin{enumerate}
    \item A hybrid difficulty scoring function combining WER with morphotonal features for curriculum learning.
    \item Gated tone conditioned adapters with 2.1M parameters that modulate encoder representations based on tonal statistics per utterance.
    \item Evaluation across Whisper, W2V-BERT and MMS on 2 datasets, revealing architecture preferences by language family.
    \item Generalisation analysis across datasets showing W2V-BERT transfers better for Nguni languages.
\end{enumerate}

\begin{figure*}[t]
\centering
\includegraphics[height=0.27\textheight]{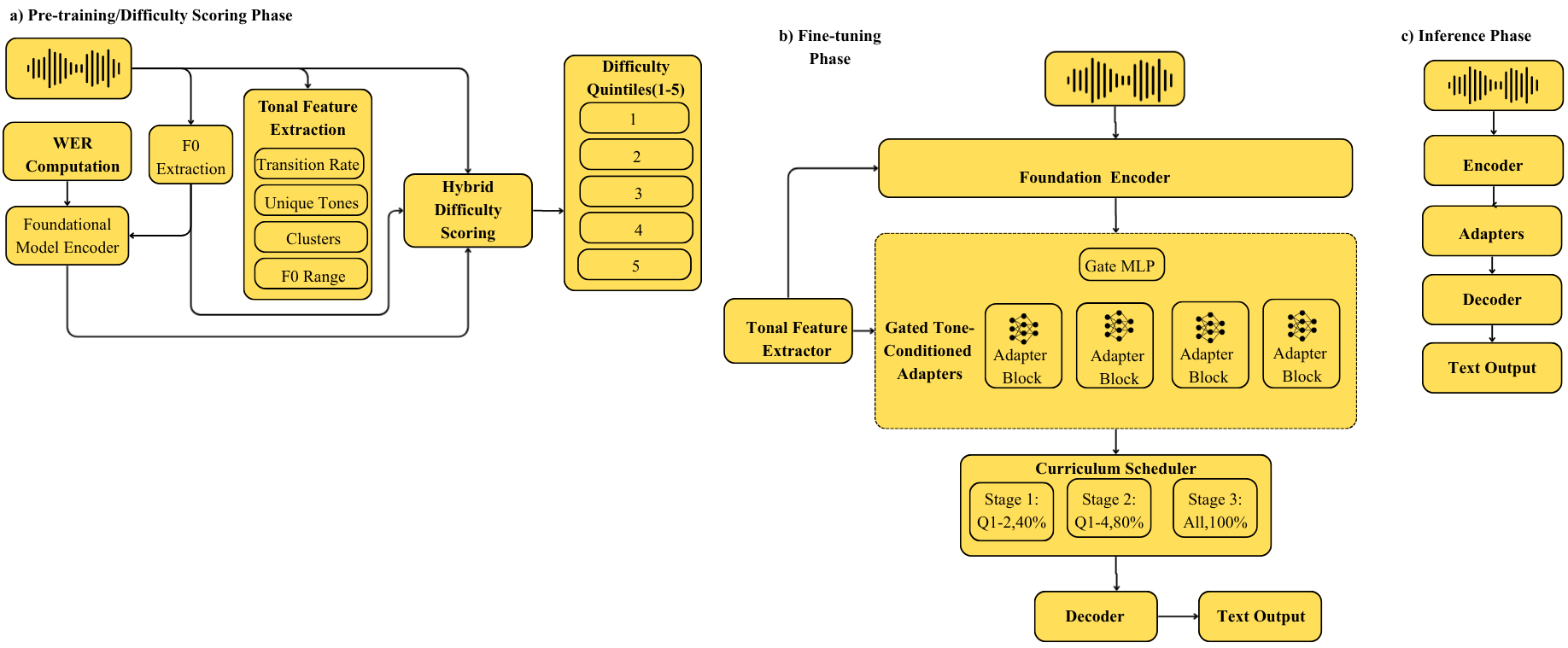}
\caption{Overview of the proposed framework. (a) Pretraining phase computes hybrid difficulty scores from WER and tonal features. (b) Fine-tuning phase uses curriculum scheduling with gated tone conditioned adapters. (c) Inference applies the trained model.}
\label{fig:framework}
\end{figure*}

\section{Related Work}
Curriculum learning in ASR orders training samples by estimated difficulty so models first learn stable patterns and then harder cases~\cite{ranjan2021curriculum, kuznetsova2022curriculum}. Difficulty has been defined by duration, signal-to-noise ratio, model loss and recognition error~\cite{karakasidis22_interspeech}. Reported gains include lower optimisation variance in low-resource settings and a 41.02\% WER reduction on dysarthric speech with curriculum plus adapters~\cite{kuznetsova2022curriculum, tan2025cba}. Hsieh et al.~\cite{hsieh2024dysarthric} further showed that difficulty inferred by the model can outperform manual ratings. In far-field ASR, Ranjan and Hansen~\cite{ranjan2021curriculum} compared CL-DH, CL-DM and CL-DHM, with CL-DM achieving up to 10.1\% relative WER reduction through progressive data merging~\cite{ranjan2017curriculum}. Karakasidis et al.~\cite{karakasidis22_interspeech} also found ordering by WER superior to scoring based on duration and loss, while related methods improved speech translation and multilingual transfer~\cite{wang2020curriculum, zhou2022exploring}.

Tone modelling remains difficult because pitch contours distinguish both word meaning and grammatical function in tonal languages~\cite{adams2017phonemic, sarma18_speechprosody}. Prior approaches either fold tone into combined vowel and tone units or model tone separately from segmental information~\cite{lee2002using, coto2021explicit}, and methods informed by pitch have improved recognition in Cantonese and Bribri, including gains of 4--25\% CER in explicit tone symbol settings~\cite{zhao21c_interspeech, coto2021explicit}. Southern Bantu tone systems differ from many East Asian systems because they involve spreading rules across word boundaries, boundary tone effects and tight coupling with morphology~\cite{mohasi2011acoustic, zerbian08_interspeech}. Mohasi et al.~\cite{mohasi2011acoustic} and Mixdorff et al.~\cite{mixdorff11_interspeech} linked these patterns to F0 behaviour and word identification, whereas Zerbian and Barnard~\cite{zerbian08_interspeech} showed measurable ASR impact. Existing studies have described the phenomenon well but rarely incorporated tonal complexity directly into curriculum design or adapter gating.

In low-resource African ASR, multilingual pretraining has improved baseline performance, yet Southern Bantu zero-shot accuracy remains weak for large models~\cite{ogunremi2023multilingual, nzeyimana2023kinspeak, zhang2023fast}, and OpenASR evaluations still report WER from 32\% to 68\% with pretrained systems~\cite{peterson2022openasr21}. This gap has both technical and sociotechnical dimensions, because dataset governance and deployment context affect usability for speaker communities~\cite{sutherland2024navajo, reitmaier2022opportunities}. Resources governed by their communities such as Swivuriso capture broader recording conditions than studio corpora such as NCHLT and therefore provide a more realistic testbed for deployment training~\cite{marivatee2025swivuriso0, badenhorst2022nchlt}, while curriculum learning with transfer across dialects and domains suggests that structured training order can improve robustness under distribution shift~\cite{suwanbandit2023thai, caubriere19_interspeech}. Our approach built on these lines by combining curriculum learning informed by WER with explicit tonal conditioning for Southern Bantu ASR.

\section{Methodology}

\begin{table*}[!htbp]
\centering
\footnotesize
\setlength{\tabcolsep}{5pt}
\begin{tabular}{l|cc|cc|cc|cc|cc|cc|cc}
\toprule
& \multicolumn{2}{c|}{\textbf{zul}} & \multicolumn{2}{c|}{\textbf{xho}} & \multicolumn{2}{c|}{\textbf{sot}} & \multicolumn{2}{c|}{\textbf{tsn}} & \multicolumn{2}{c|}{\textbf{ven}} & \multicolumn{2}{c|}{\textbf{tso}} & \multicolumn{2}{c}{\textbf{Avg}} \\
\textbf{Model} & S & N & S & N & S & N & S & N & S & N & S & N & S & N \\
\midrule
\multicolumn{15}{l}{\textit{Whisper-based Models}} \\
Whisper Multilingual & \textbf{28.12} & 39.48 & \textbf{30.73} & 43.93 & \textbf{23.30} & \textbf{31.81} & 19.24 & \textbf{28.07} & 17.27 & 39.50 & \textbf{21.37} & 30.38 & \textbf{23.34} & \textbf{35.53} \\
Whisper Tone-cond. & 28.59 & \textbf{38.44} & 30.39 & \textbf{43.12} & 23.82 & 36.39 & 18.74 & 36.52 & \textbf{17.23} & 41.67 & 23.26 & 38.15 & 23.67 & 39.05 \\
Whisper Tone+Curr. & 29.10 & 39.81 & 31.38 & 44.46 & 23.53 & 36.26 & \textbf{18.60} & 30.19 & 17.50 & 45.79 & 21.67 & 33.56 & 23.63 & 38.34 \\
Whisper Curriculum & 29.23 & 42.39 & 31.61 & 43.95 & 23.86 & 33.62 & 18.78 & 27.54 & 18.98 & 42.91 & 21.55 & \textbf{31.80} & 24.00 & 37.04 \\
\midrule
\multicolumn{15}{l}{\textit{W2V-BERT Models}} \\
W2V-BERT Multilingual & 27.16 & 38.69 & 29.00 & 47.34 & 27.66 & 34.67 & 22.79 & 33.55 & \textbf{20.25} & \textbf{32.81} & 27.03 & 29.71 & 25.65 & 36.13 \\
W2V-BERT Tone-cond. & 24.99 & 35.37 & 26.02 & 43.76 & \textbf{24.53} & \textbf{31.15} & \textbf{19.84} & \textbf{28.43} & 25.93 & 35.52 & \textbf{21.54} & \textbf{23.79} & \textbf{23.81} & \textbf{33.00} \\
W2V-BERT Tone+Curr. & \textbf{24.79} & \textbf{34.63} & \textbf{26.06} & \textbf{40.77} & 25.23 & 37.42 & 20.92 & 34.59 & 26.61 & 39.28 & 25.17 & 26.32 & 24.80 & 35.50 \\
W2V-BERT Curr.  & 24.88 & 36.49 & 26.44 & 41.33 & 28.55 & 31.96 & 23.88 & 30.27 & 26.65 & 39.27 & 24.54 & 24.46 & 25.82 & 33.96 \\
\midrule
\multicolumn{15}{l}{\textit{MMS-based Models}} \\
MMS Multilingual & 35.19 & 35.77 & 38.65 & 46.87 & 45.29 & 50.61 & 40.65 & 46.19 & 45.31 & 52.98 & 41.02 & 41.50 & 41.02 & 45.65 \\
MMS Tone-cond. & 35.10 & 42.97 & 38.47 & 62.51 & 49.47 & 62.62 & 42.09 & 55.36 & 51.22 & 63.13 & 50.99 & 52.22 & 44.56 & 56.47 \\
MMS Tone+Curr. & \textbf{34.82} & \textbf{41.27} & 38.60 & 59.72 & 47.61 & 63.35 & 41.99 & 55.78 & 51.12 & 62.95 & 54.17 & 54.19 & 44.72 & 56.21 \\
MMS Curriculum & 33.60 & 35.66 & \textbf{36.79} & \textbf{43.38} & \textbf{41.11} & \textbf{46.59} & \textbf{37.14} & \textbf{46.56} & \textbf{39.68} & \textbf{50.61} & \textbf{36.53} & \textbf{42.66} & \textbf{37.48} & \textbf{44.24} \\
\bottomrule
\end{tabular}
\caption{Per-language WER (\%) on Swivuriso (S) and NCHLT (N). Best per architecture in \textbf{bold}. Configurations are Multilingual (joint multilingual fine-tuning without tone adapters or curriculum), Tone-cond. (tone conditioned adapters), Tone+Curr. (tone conditioned adapters with hybrid curriculum, WER+tonal) and Curriculum (WER only curriculum without tone adapters).}
\label{tab:per_lang}
\end{table*}

Figure~\ref{fig:framework} presents the tone conditioned curriculum framework, guided by 2 constraints, namely limited Southern Bantu training data and preservation of tonal linguistic structure. The pipeline comprised 3 phases, namely hybrid difficulty scoring, staged fine-tuning with gated adapters and inference.

\subsection{Hybrid Difficulty Scoring}

To formalise curriculum difficulty, we defined a hybrid score that balanced empirical model performance and linguistic complexity. Following Karakasidis et al.~\cite{karakasidis22_interspeech}, we used scoring by WER as the primary component and computed each utterance score $s(u)$ as:
\begin{equation}
    s(u) = \alpha \cdot \text{WER}_{\text{norm}}(u) + \beta \cdot \text{Tonal}_{\text{norm}}(u)
\end{equation}
where $\alpha=0.7$ and $\beta=0.3$ weight WER and tonal components. We set $\alpha>\beta$ because recognition error proved the strongest single difficulty indicator~\cite{karakasidis22_interspeech}, and we leave sensitivity analysis to future work. $\text{WER}_{\text{norm}}$ was computed using a frozen fine-tuned Whisper baseline trained on Swivuriso, and all scores were precomputed before curriculum training. $\text{Tonal}_{\text{norm}}$ aggregated morphotonal features, with higher weights on transition density (0.3) and unique pattern variety (0.2), and we then mapped utterances to difficulty quintiles for staged pacing.

\subsection{Tonal Feature Extraction}

Our feature pipeline quantified tonal complexity per utterance without forced alignment or manual phonetic transcription and thus avoided dependence on scarce expert annotation. We extracted F0 contours at 10ms intervals using Parselmouth~\cite{mohasi2011acoustic} with a 75--500Hz pitch range and, after filtering unvoiced frames, computed 5 features, namely transition rate, unique tone count, tone cluster count, F0 standard deviation and F0 range. Features were z-score normalised per language using training set statistics, and because pitch carries both meaning and grammatical distinctions in these languages, higher transition rates correlated with higher recognition difficulty. We discretised F0 into semitones relative to the utterance mean and binned values into 5 tone levels (High-High, High, Mid, Low, Low-Low) with thresholds at $\pm$2 and $\pm$6 semitones~\cite{zerbian08_interspeech}, thereby capturing relative tonal targets independently of absolute pitch.

\subsection{Tone-Conditioned Gated Adapters}

We injected tonal context at the utterance level into the encoder through 4 parallel gated bottleneck adapters applied after the final encoder layer and before the decoder. A 2-layer gate MLP mapped the 5-dimensional tonal vector $\mathbf{f}_{\text{tone}}$ to gate values $\mathbf{g} = \sigma(\text{MLP}(\mathbf{f}_{\text{tone}}))$, using a 128-unit hidden layer to map the 5 tonal inputs to 4 gate values, with ReLU and sigmoid activations. The final encoder output $\mathbf{y}$ is
\begin{equation}
    \mathbf{y} = \mathbf{x} + \sum_{i=1}^{N} g_i \cdot \mathbf{a}_i(\text{LN}(\mathbf{x}))
\end{equation}
where $\mathbf{a}_i$ are bottleneck linear layers that project from the hidden size down to 256 and back up to the hidden size. The adapters added approximately 2.1M parameters (0.3\% of Whisper-large-v3-turbo) with shared gate weights across adapters, and this lightweight design was therefore trainable on small community datasets without retraining a full foundation model. When tonal cues were informative, the gate values increased adapter contribution, whereas for utterances with simpler tonal profiles the gate reduced adapter influence and retained pretrained representations.

\subsection{Staged Curriculum Training}

We implemented a schedule with 3 stages inspired by CL-DM~\cite{ranjan2021curriculum}. With limited training data, exposing the full difficulty range at once risked overfitting the hardest, noisiest samples before core patterns were learnt, so we introduced harder quintiles progressively with fixed step allocation. Stage~1 at steps 1--650 used the easiest 40\% of samples, Stage~2 at steps 651--1300 expanded to the easiest 80\%, and Stage~3 at steps 1301--2000 used the full training set. Fixed pacing improved reproducibility, and progressive merging helped preserve performance on easier structures while adapting to harder tonal patterns.

\section{Experiments}

\subsection{Dataset}

We used the Swivuriso/za-african-next-voices dataset~\cite{marivatee2025swivuriso0}, a corpus developed by speaker communities with read speech from 6 Southern Bantu languages; isiZulu with 1,810 samples, isiXhosa with 2,470, Sesotho with 2,717, Setswana with 4,299, Tshivenda with 1,192 and Xitsonga with 3,500. The dev\_test split contained 15,988 samples across 26.7 hours. The corpus is governed by its speaker communities, and for evaluation across datasets we used NCHLT with 41,964 samples across the same 6 languages.

Table~\ref{tab:difficulty} reports training set difficulty statistics. isiZulu and isiXhosa had the highest mean difficulty scores at 0.39, whereas Setswana had the lowest at 0.29. The tonal column reports normalised morphotonal complexity, where negative values indicate below average and positive values above average complexity, and Xitsonga had the highest WER standard deviation at 0.508 and therefore showed high variability in sample difficulty. Audio was resampled to 16kHz, utterances outside 0.5--30 seconds were removed, and text was normalised by lowercasing, removing noise tags such as "?" and "cs" markers, stripping punctuation and diacritics and normalising whitespace.

\begin{table}[H]
\centering
\begin{tabular}{lccccc}
\toprule
\textbf{Lang} & \textbf{N} & \textbf{WER} & \textbf{$\sigma$WER} & \textbf{Diff.} & \textbf{Tonal} \\
\midrule
zul & 4,624 & 0.350 & 0.244 & 0.388 & $-$0.044 \\
xho & 4,833 & 0.344 & 0.226 & 0.386 & 0.036 \\
sot & 4,758 & 0.269 & 0.383 & 0.321 & $-$0.008 \\
tsn & 4,939 & 0.201 & 0.308 & 0.292 & $-$0.011 \\
ven & 4,954 & 0.235 & 0.336 & 0.314 & 0.028 \\
tso & 4,942 & 0.232 & 0.508 & 0.270 & $-$0.001 \\
\bottomrule
\end{tabular}
\caption{Training set difficulty statistics by language. N is the number of training utterances per language. WER and $\sigma$WER are the mean and standard deviation of per-utterance WER from the frozen Whisper baseline. Diff. is the mean hybrid difficulty score and Tonal is the normalised morphotonal complexity.}
\label{tab:difficulty}
\end{table}

\subsection{Training and Evaluation}

Models were trained on 2 NVIDIA A100 80GB GPUs with bfloat16 precision, AdamW ($\beta_2=0.98$), gradient clipping at 1.0, and seed 42. For W2V-BERT, we used an effective batch size of 32 via 8$\times$4 gradient accumulation, a learning rate of $5\times10^{-5}$, and 500 warmup steps, whereas Whisper used an effective batch size of 32 via 2$\times$16 accumulation, a learning rate of $1\times10^{-5}$, and 100--500 warmup steps depending on configuration. Training ran for 2,000 steps, and curriculum variants used stages of 650 steps each. For W2V-BERT CTC training, we built a unified multilingual vocabulary of 80--100 tokens with interleaved language sampling and set \texttt{ctc\_zero\_infinity=True} for stability. Dropout was disabled in all runs, and datasets were streamed from HuggingFace Hub to reduce disk usage. Evaluation used greedy decoding for consistency across CTC and seq2seq models, and we reported Word Error Rate, Character Error Rate and BERTScore F1 with \texttt{bert-base-multilingual-cased}. All models were trained only on Swivuriso and evaluated on Swivuriso dev\_test (matched) and NCHLT (transfer), without adaptation or hyperparameter retuning. We evaluated 4 configurations per architecture, namely multilingual baseline fine-tuning, Tone-cond. (gated tone adapters only), Tone+Curr. (tone adapters with hybrid curriculum, $\alpha$=0.7 and $\beta$=0.3) and Curriculum (ordering by WER alone without tone adapters).

\subsection{Results and Discussion}

Table~\ref{tab:main_results} reports combined averages across Swivuriso and NCHLT. Before adaptation, Whisper and MMS both failed on these languages, consistent with Reitmaier et al.~\cite{reitmaier2022opportunities}, but fine-tuning reduced error rates substantially. On Swivuriso alone in Table~\ref{tab:per_lang}, Whisper Multilingual reached 23.34\% WER, an 83.6\% relative reduction from the 142.73\% zero-shot baseline, and the higher values in Table~\ref{tab:main_results}, for example 29.44\% for Whisper Multilingual, reflect the inclusion of NCHLT. Table~\ref{tab:per_lang} also reveals strong variation across languages that averaged metrics can obscure. For Nguni languages (isiZulu and isiXhosa), W2V-BERT yielded lower WER than Whisper, with isiZulu at 24.79\% versus 28.12\% on Swivuriso and 34.63\% versus 39.48\% on NCHLT, and isiXhosa at 26.06\% versus 30.73\%. Sotho-Tswana languages showed the opposite pattern, where Whisper outperformed W2V-BERT and Sesotho reached 23.30\% with Whisper Multilingual versus 24.53\% for the best W2V-BERT setting, and Setswana recorded the lowest WER in the study with Whisper Tone+Curr.\ at 18.60\% on Swivuriso and 27.54\% on NCHLT.

\begin{table}[H]
\centering
\begin{tabular}{lccc}
\toprule
\textbf{Model} & \textbf{WER}$\downarrow$ & \textbf{CER}$\downarrow$ & \textbf{BERT}$\uparrow$ \\
\midrule
\multicolumn{4}{l}{\textit{Zero-shot Baselines}} \\
Whisper Large-v3-Turbo & 146.30 & 71.66 & 0.676 \\
MMS-1B-All & 112.98 & 85.01 & 0.642 \\
\midrule
\multicolumn{4}{l}{\textit{Whisper-based Models}} \\
Whisper Multilingual & \textbf{29.44} & \textbf{7.44} & \textbf{0.936} \\
Whisper Tone-cond. & 31.36 & 8.87 & 0.934 \\
Whisper Tone+Curr. & 30.99 & 8.91 & 0.933 \\
Whisper Curriculum & 30.52 & 7.85 & 0.934 \\
\midrule
\multicolumn{4}{l}{\textit{W2V-BERT Models}} \\
W2V-BERT Multilingual & 30.89 & \textbf{6.83} & 0.932 \\
W2V-BERT Tone-cond. & \textbf{28.41} & 6.98 & \textbf{0.934} \\
W2V-BERT Tone+Curr. & 29.38 & 7.25 & 0.931 \\
W2V-BERT Curr. & 30.66 & 7.46 & 0.928 \\
\midrule
\multicolumn{4}{l}{\textit{MMS-based Models}} \\
MMS Multilingual & 43.34 & 10.25 & 0.899 \\
MMS Tone-cond. & 50.52 & 12.30 & 0.880 \\
MMS Tone+Curr. & 50.47 & 12.42 & 0.879 \\
MMS Curriculum & \textbf{40.86} & \textbf{9.72} & \textbf{0.905} \\
\bottomrule
\end{tabular}
\caption{Combined average ASR performance on Swivuriso and NCHLT. Best per architecture in \textbf{bold}. WER and CER are percentages. See Table~\ref{tab:per_lang} for per-dataset values.}
\label{tab:main_results}
\end{table}

Tshivenda attained the lowest matched WER at 17.23\% with Whisper Tone-cond.\ on Swivuriso but also showed the largest transfer increase, rising to 39.50\% on NCHLT, a 22 point gap consistent with domain mismatch between community recorded Swivuriso and studio recorded NCHLT. W2V-BERT Multilingual recorded lower NCHLT WER for Tshivenda at 32.81\%, and Xitsonga showed stronger transfer stability with W2V-BERT Tone conditioned, moving from 21.54\% on Swivuriso to 23.79\% on NCHLT, whereas Whisper increased by approximately 9 points from 21.37\% to 30.38\%. Across model families, tone conditioning was dependent on architecture and curriculum effects were mixed. Whisper gains were small at average WER 23.34 to 23.67\%, whereas W2V-BERT improved from 25.65\% to 23.81\%, a 7.2\% relative gain that indicated stronger benefit in CTC decoding. Hybrid curriculum did not transfer uniformly because Whisper curriculum variants fell slightly below baseline, MMS performed best with curriculum based on WER alone, and W2V-BERT Tone+Curr.\ did not exceed Tone-cond.\ alone. Curriculum effects also varied by language, with gains for isiZulu and Setswana but degradation for Tshivenda under W2V-BERT Tone+Curr., so model choice per language and validation across datasets remained necessary before deployment.

\section{Conclusions}

We presented a tone conditioned curriculum framework for 6 Southern Bantu languages and evaluated 12 configurations across Whisper, W2V-BERT and MMS on both Swivuriso and NCHLT. The central finding was that architecture suitability varied by language family, with W2V-BERT achieving lower error rates on Nguni languages whereas Whisper held an advantage on Sotho-Tswana languages, a split that global averages masked. Tone conditioning reinforced this divergence, reducing W2V-BERT error by 7.2\% relative but offering little benefit to the other 2 architectures, which suggested that the gated adapters were most effective with CTC decoding. Curriculum scheduling produced uneven outcomes, improving some languages but degrading others, and for MMS the simpler ordering by WER alone proved more effective than the hybrid variant. Generalisation across corpora depended on both language and architecture, as some pairings transferred well while others degraded by over 20~WER points. No single model sufficed for all 6 languages, and reliable deployment therefore requires selecting architectures per language and validating across recording conditions. The framework also showed that tonal structure can inform parameter-efficient adaptation without manual phonological annotation. Future work will evaluate unseen conversational and code-switched speech as well as test transfer to other tonal African language families.

\section{Acknowledgements}
 We thank all collaborators, institutions, and partners who contributed to the African Next Voices project (funded through a grant from Gates Foundation and a gift from Meta). We acknowledge the support and engagement of Masakhane, Lanfrica, the University of Pretoria (ITS and Research Office). The authors acknowledge the Data Science for Social Impact (DSFSI) Lab, supported by the ABSA Chair of Data Science and the African Institute of Data Science and AI (AfriDSAI). This work was supported by the AI4D Africa Program (African Languages Lab), funded by the International Development Research Centre (IDRC), Canada and Foreign, Commonwealth \& Development Office (FCDO), UK. 
 We acknowledge the support of the  Research Ireland Centre for Research Training in Digitally Enhanced Reality (d-real), Grant Nos. 18/CRT/6224 and 19/FFP/6917, and by the ADAPT Research Ireland Centre for AI-Driven Digital Content Technology, Grant No. 13/RC/2106-P2.

\section{Generative AI Use Disclosure}
Generative AI tools were used solely to assist with language editing and clarity of presentation. All research
ideas, methodology, experiments as well as interpretations were conceived and carried out by the authors, who take full responsibility for the originality, validity, and integrity of the work.

\bibliographystyle{IEEEtran}
\bibliography{mybib}

\end{document}